\pdfoutput=1
\documentclass[conference]{IEEEtran}
\IEEEoverridecommandlockouts
\usepackage{algorithmic}
\usepackage{amsmath,amssymb,amsfonts}
\usepackage{booktabs} 
\usepackage{cite}
\usepackage{cleveref}
\crefname{appendix}{Appendix}{Appendices}
\Crefname{appendix}{Appendix}{Appendices}
\crefname{section}{Section}{Sections}
\Crefname{section}{Section}{Sections}
\usepackage{graphicx}
\usepackage{textcomp}
\usepackage{xcolor}

\usepackage[table]{xcolor}
\definecolor{highlightcolor}{HTML}{EAF4F5}

\def\BibTeX{{\rm B\kern-.05em{\sc i\kern-.025em b}\kern-.08em
    T\kern-.1667em\lower.7ex\hbox{E}\kern-.125emX}}

\begin{document}
\title{Knowledge-Refined Dual Context-Aware Network for Partially Relevant Video Retrieval}
\author{
    Junkai Yang$^{1,\dagger}$, Qirui Wang$^{1,\dagger}$, Yaoqing Jin$^{2,\dagger}$, Shuai Ma$^{1}$, Minghan Xu$^{1}$, Shanmin Pang$^{1,\ast}$ \\
    $^{1}$School of Software Engineering, Xi'an Jiaotong University \\
    $^{2}$Faculty of Computer Science, Electrical Engineering and Information Technology, Universität Stuttgart \\
    \ \\
    $^{\dagger}$ These authors contributed equally to this work. \\
    $^{\ast}$ Corresponding author.
}

\maketitle

\section{Abstract}
\label{sec:abstract}

Retrieving partially relevant segments from untrimmed videos remains difficult due to two persistent challenges: the mismatch in information density between text and video segments, and limited attention mechanisms that overlook semantic focus and event correlations. We present KDC-Net, a Knowledge-Refined Dual Context-Aware Network that tackles these issues from both textual and visual perspectives. On the text side, a Hierarchical Semantic Aggregation module captures and adaptively fuses multi-scale phrase cues to enrich query semantics. On the video side, a Dynamic Temporal Attention mechanism employs relative positional encoding and adaptive temporal windows to highlight key events with local temporal coherence. Additionally, a dynamic CLIP-based distillation strategy, enhanced with temporal-continuity-aware refinement, ensures segment-aware and objective-aligned knowledge transfer. Experiments on PRVR benchmarks show that KDC-Net consistently outperforms state-of-the-art methods, especially under low moment-to-video ratios.

\begin{IEEEkeywords}
Relevant Video Retrieval, Video Analysis, Cross-model Matching
\end{IEEEkeywords}

\section{Introduction}
\label{sec:intro}
With the rapid rise of short-video platforms and video-centric social media \cite{fei2024video}, efficiently retrieving relevant content from massive collections of untrimmed videos has become increasingly crucial. To address this need, the task of Partially Relevant Video Retrieval (PRVR) has emerged \cite{dong2022partially}\cite{wang2024gmmformer}\cite{yin2024exploiting}, aiming to localize segments within raw videos that partially match a given textual query. Unlike traditional text-to-video retrieval (T2VR) \cite{chen2023joint}, which operates on pre-trimmed clips \cite{cho2025ambiguity}\cite{wang2022cross}, PRVR more faithfully reflects real-world scenarios and has gained considerable attention in recent years.
\begin{figure}[!t]
 \begin{center}
    \includegraphics[width=1\linewidth]{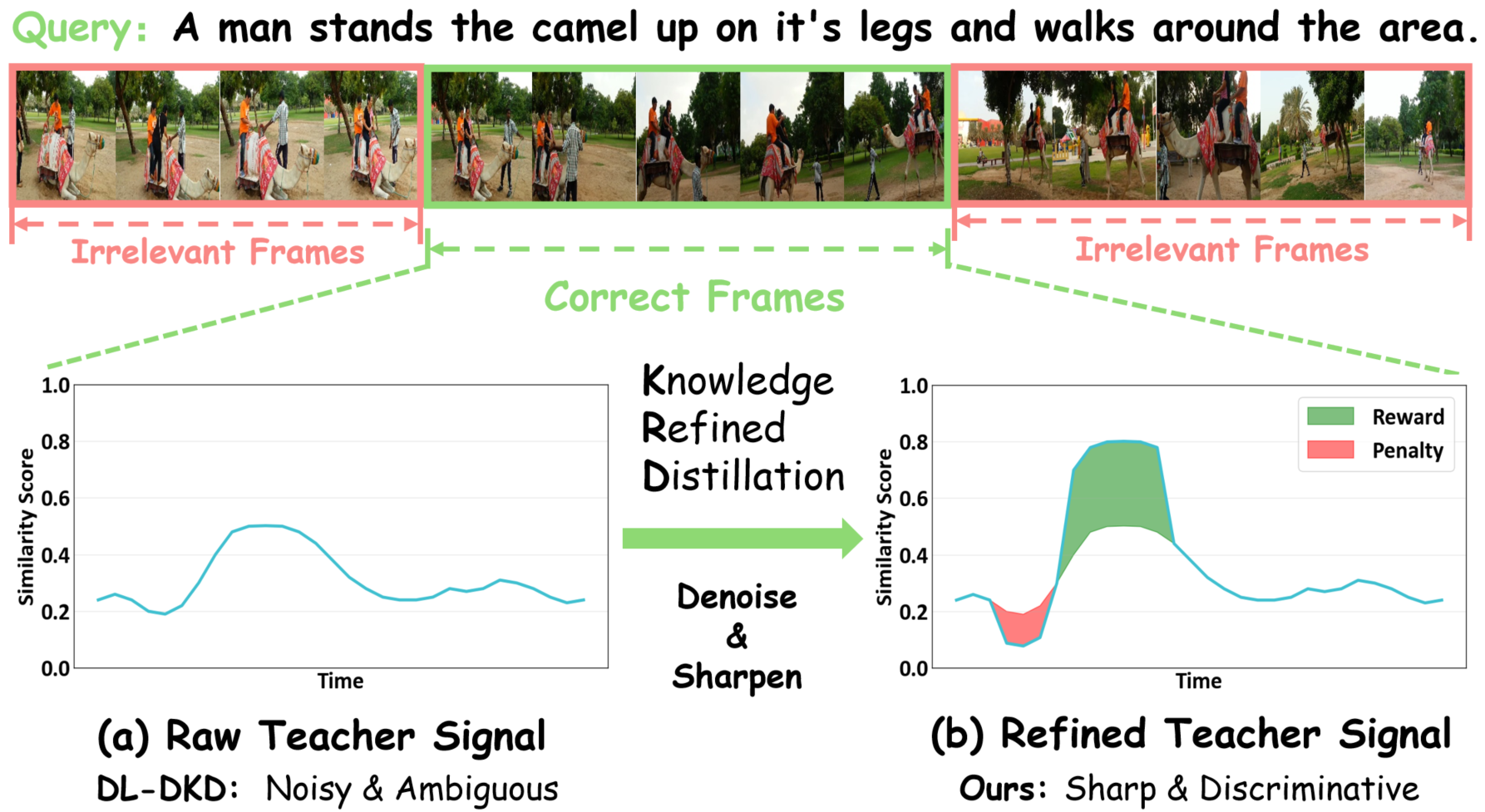} 
    \caption{The Overview diagram illustrates the task objectives of PRVR, and also shows that the Knowledge Refined Distillation strategy we proposed can effectively optimize the distilled signals obtained from the teacher model.}
    \label{fig:KDR_Ablation}
  \end{center}
\end{figure}

Despite notable progress, existing approaches still face two fundamental challenges. First, on the textual side, most current models rely on Transformer-based flattened self-attention, which lacks explicit modeling of the inherent local compositional structure of natural language \cite{wangEfficient2025}. As a result, core semantic units—namely multi-word phrases—tend to be diluted \cite{wang2024gmmv2} \cite{wang2026personalq}, preventing the model from forming a properly hierarchical understanding of complex queries. Second, on the visual side, directly applying standard self-attention to frame sequences introduces a form of temporal agnosticism: permutation invariance prevents the model from inherently capturing event-level temporal dependencies \cite{ren2025exploiting}, while the global and uniform attention computation ignores the strong locality priors essential for segment retrieval. This not only leads to inefficient computation but also makes the model highly susceptible to noise from irrelevant frames. Furthermore, many state-of-the-art methods rely on knowledge distillation from large pretrained vision-language models such as CLIP. Yet, standard distillation follows a blind imitation paradigm since CLIP is not tailored for video understanding \cite{zhang2025multi}, and its similarity distributions often contain noise and ambiguity \cite{croitoru2025teachtext}. A student model that imitates these signals uncritically wastes capacity trying to fit imperfect supervision, ultimately limiting its performance ceiling.

To address these challenges systematically, we propose the Knowledge-Refined Dual Context-Aware Network (KDC-Net), which introduces coordinated innovations across text representation, video modeling, and training methodology.

To resolve the lack of hierarchical textual information, we design the Hierarchical Semantic Aggregation (HSA) module. It explicitly extracts multi-scale textual phrases using sliding windows and applies self-attention within each phrase to capture local dependencies, producing high-quality phrase-level features. A learnable fusion mechanism then adaptively integrates these multi-scale phrase representations back into the token sequence, yielding a dense, compositionally enriched, and hierarchically structured query representation.

To address the sparsity of semantic focus in video modeling, we introduce the Dynamic Temporal Attention (DTA) mechanism. DTA injects learnable relative positional biases to capture temporal relationships more effectively and employs dynamically generated multi-scale temporal windows to restrict the attention field. This encourages the model to focus computational resources on temporally coherent key events while suppressing irrelevant noise, enabling robust modeling of both short-term motions and long-term dynamics.

Finally, to overcome the limitations of conventional distillation, we propose a novel Knowledge Refinement Distillation (KRD) strategy. Instead of directly using CLIP’s similarity distribution as supervision, KRD acts as a critical evaluator, Leveraging the prior that real events exhibit temporal continuity, we identify consecutively high-confidence or low-confidence regions in the teacher distribution via dynamic thresholding. High-confidence segments are amplified, low-confidence ones are suppressed, and the refined, cleaner distribution serves as the student’s learning target. This refinement substantially improves the quality and utility of the distillation signal, leading to more efficient and robust knowledge transfer.

In summary, our main contributions are as follows:
\begin{itemize}
\item We introduce the HSA module, which explicitly models and adaptively fuses phrase-level semantics to produce richly structured textual representations.
\item We develop the DTA mechanism, which injects relative positional priors and dynamic temporal windows into attention computation, enabling accurate and efficient modeling of multi-scale temporal dynamics.
\item We propose the KRD strategy, which denoises and enhances teacher signals prior to distillation, enabling more robust and effective knowledge transfer. Extensive experiments across multiple benchmarks demonstrate the superiority of our method.
\end{itemize}

\section{Related Work}
\label{sec:Related Work}

\subsection{Partially Relevant Video Retrieval}
PRVR aims to locate segments in untrimmed videos matching a textual query, demanding both cross-modal alignment and temporal localization. Recent methods have explored diverse strategies: MS-SL \cite{dong2022partially} integrates multi-level features, PEAN \cite{jiang2023progressive} uses hierarchical modeling, T-D3N \cite{cheng2024transferable} enhances textual semantics, and GMMFormer \cite{wang2024gmmformer} models frame dependencies with Gaussian constraints. DL-DKD \cite{dong2023dual} leverage large-scale pretraining or distill knowledge from CLIP. However, these methods often rely on flat attention mechanisms that neglect the hierarchical semantics of text queries. Furthermore, their video modeling struggles to capture multi-scale temporal dynamics and is susceptible to noise from irrelevant content. In contrast, our work addresses these gaps with a hierarchical text encoder and a dynamic temporal attention mechanism for video.

\subsection{Knowledge Distillation}
Knowledge distillation is widely used in cross-modal retrieval to transfer the alignment capabilities of large vision-language models like CLIP to smaller student models \cite{hinton2015distilling} \cite{meng2025evdclip} \cite{wang2024internvideo2}. For instance, TeachText applies knowledge distillation to text-video retrieval \cite{radford2021learning}, GCKD \cite{luo2025graph} introduces the teacher model to generate cross-domain pseudo-similarity labels to alleviate the distribution shift problem in cross-domain transfer, DL-DKD \cite{dong2023dual} uses a dual-branch framework. However, these approaches often employ blind imitation paradigm, treating the teacher's output as a perfect target. This overlooks inherent noise, especially when applying image-text models like CLIP to video, forcing the student to fit suboptimal signals. Our proposed Knowledge Refinement Distillation (KRD) overcomes this by first refining the teacher's noisy signal using temporal continuity priors before guiding the student, enabling more robust knowledge transfer.

\section{Methodology}
\label{sec:method}

\begin{figure*}[ht]
    \centering
    \includegraphics[width=1.0\linewidth]{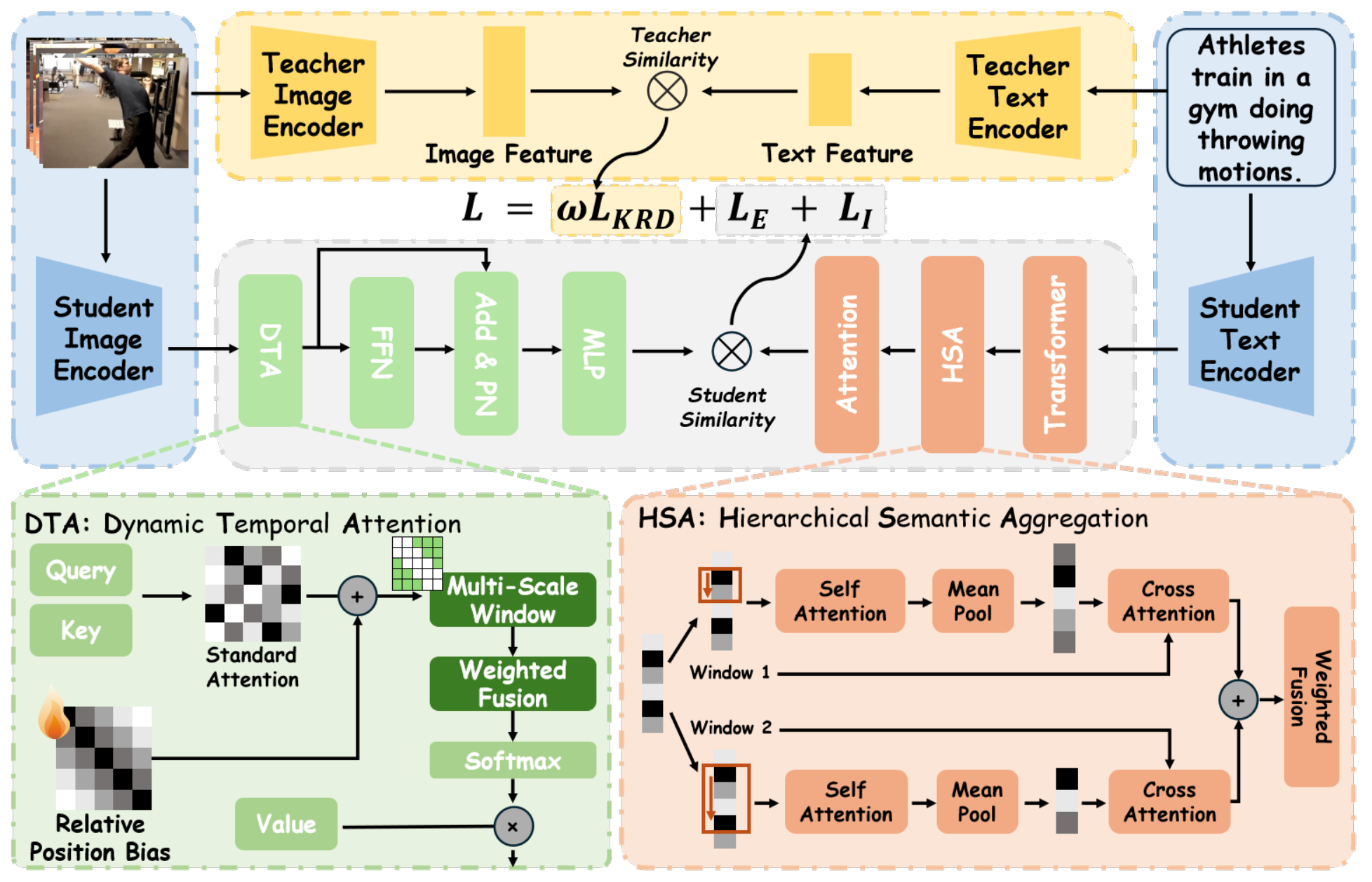} 
    \caption{Illustration of KDC-Net. It employs a distillation framework, the student model comprises two independent branches with no parameter sharing.}
    \label{fig:KDC-Net-PipeLine}
\end{figure*}

KDC-Net is designed to learn robust and fine-grained representations for both text and video to address the core challenges in PRVR. It consists of three key components: a Hierarchical Semantic Aggregation (HSA) module for text encoding, a Dynamic Temporal Attention (DTA) mechanism for video modeling, and a novel Knowledge Refinement Distillation (KRD) strategy for training.

\subsection{Hierarchical Semantic Aggregation for Text}
To capture the compositional nature of language, we introduce the HSA module. Given word-level features $\mathbf{Q} = \{\mathbf{q}_i\}_{i=1}^L$, HSA first employs multi-scale sliding windows to generate a set of phrase-level features $\mathbf{P}^{(m)}$ for each scale $m \in \mathcal{M}$. Specifically, each phrase feature is extracted by applying Multi-Head Attention followed by average pooling within its corresponding window to model local context.

Then, to enrich the original word features with hierarchical context, each word feature $\mathbf{q}_i$ is updated via a two-stage aggregation. First, for each scale $m$, a scale-specific context vector $\mathbf{a}_i^{(m)}$ is computed by attending to all phrase features within that scale:
\begin{equation}
\mathbf{a}_i^{(m)} = \sum_{j} \text{softmax}(\frac{\mathbf{q}_i \cdot \mathbf{p}^{(m)}_j}{\sqrt{D}}) \mathbf{p}^{(m)}_j
\end{equation}
where $\mathbf{p}^{(m)}_j \in \mathbf{P}^{(m)}$. This allows each word to aggregate information from semantically relevant phrases at a specific granularity. Second, these scale-specific vectors are fused using learnable importance weights $\beta^{(m)}$ to form the final enhanced representation:
\begin{equation}
\mathbf{q}'_i = \text{LayerNorm}(\mathbf{q}_i + \sum_{m \in \mathcal{M}} \beta^{(m)} \mathbf{a}_i^{(m)})
\end{equation}
Finally, pooling the enhanced word features $\{\mathbf{q}'_i\}_{i=1}^L$ to produces the final query representation $\mathbf{q} \in \mathbb{R}^D$.

\subsection{Dynamic Temporal Attention for Video}
For a sequence of video frame features $\mathbf{V} = \{\mathbf{v}_i\}_{i=1}^K \in \mathbb{R}^{K \times D}$, we propose the DTA mechanism to replace the standard self-attention in our video encoder. DTA injects two critical temporal priors into the attention calculation to effectively model temporal dynamics and suppress noise. The attention score between a query frame $i$ and a key frame $j$ is formulated as:
\begin{equation}
\text{score}(i, j) = \frac{\mathbf{q}_i^\top \mathbf{k}_j}{\sqrt{d_k}} + b_{i,j} + m_{i,j}
\end{equation}
where the first term is the standard dot-product attention. The second term, $b_{i,j}$, is a learnable relative position bias. It explicitly encodes temporal ordering by mapping the relative distance $(j-i)$ to a learnable embedding, allowing the model to distinguish frames based on their temporal separation. The third term, $m_{i,j}$, is a dynamic window mask that restricts attention to a local neighborhood around frame $i$, defined as:
\begin{equation}
m_{i,j} = 
\begin{cases} 
0, & \text{if } |j-i| \le U \\
-\infty, & \text{otherwise}
\end{cases}
\end{equation}
where $U$ is the window radius. By employing multiple window sizes, this mask focuses computation on temporally relevant frames and adapts to different action scales. This design allows DTA to efficiently capture both short-term actions and long-term dependencies.

Furthermore, to address the high temporal redundancy in video features, we introduce Purification Normalization (PN) as a lightweight substitute for standard LayerNorm. PN explicitly purifies features by identifying and removing redundant information. It computes a self-similarity matrix $S$ from the input features $x$ to model intra-sequence redundancy, then subtracts the redundancy-focused representation from the original features. The process is formulated as:
\begin{equation}
x_{\text{out}} = \text{LayerNorm}(x - \lambda \cdot (S \cdot x))
\end{equation}
where $\lambda$ is a scale factor, and $S = \text{Softmax}(\hat{x} \cdot \hat{x}^\top)$ with $\hat{x}$ being the L2-normalized features. This allows the video encoder to produce more discriminative representations.

\subsection{Knowledge Refinement Distillation}
Given a query-video pair \((Q, V)\), we utilize CLIP's image and text encoders to extract modality-specific features. Specifically, each frame $v_i \in V$ is encoded into a visual embedding $f_i^t$, yielding the frame-level representation \(F^t = \{f_i^t\}_{i=1}^K\in \mathbb{R}^{K \times d}\), while the query $Q$ is mapped to a global textual feature \(q^t \in \mathbb{R}^{1 \times d}\), where \(d\) is the shared embedding dimension.

To effectively transfer the semantic alignment capability of the teacher model into the student network, we focus on distilling the similarity distribution between the query and individual video frames. Formally, the semantic similarity distribution \(S^t \in \mathbb{R}^K\) is computed as:
\begin{equation}
S^t = [\cos(f_1^t, q^t), \cos(f_2^t, q^t), ..., \cos(f_K^t, q^t)]
\end{equation}

To leverage the power of large pre-trained models like CLIP while avoiding the blind imitation pitfall of standard Knowledge Distillation, we propose a novel KRD strategy. Confidence thresholds are dynamically determined based on the current batch's prediction statistics. Specifically, given a teacher-generated score sequence \(S^t = \{s_1^t, s_2^t, ..., s_K^t\}\),  we compute the mean \(\mu_s=\frac{1}{K}\sum_{i=1}^{K}s_i^t\) and standard deviation \(\sigma_s=\sqrt{\frac{1}{K}\sum_{i=1}^{K}(s_i^t - \mu_s)^2}\).
Using these statistics, we set adaptive thresholds as:
\begin{equation}
\tau_{high} = \mu_s + \sigma_s,\quad  
\tau_{low} = \mu_s - \sigma_s
\end{equation}
which allow flexible boundary adjustment to accommodate varying confidence distributions across video sequences.

To capture temporal continuity, we employ a sliding-window local decision strategy that models confidence correlations among adjacent frames. For each score \( s_i^t \), a local window \( w_i = [s_i^t, s_{i+1}^t, \ldots, s_{i+k}^t] \) with window size \( k \) is defined. If all values in \( w_i \) exceed the high threshold \( \tau_{high} \), the region is identified as high-confidence and enhanced; if all fall below the low threshold \( \tau_{low} \), it is penalized. The adjusted score is computed as:
\begin{equation}
\tilde{s}_i^t = s_i^t + I(i) \cdot \alpha,\quad
I(i) =
\begin{cases}
1, & \text{if } W_i \geq \tau_{\text{high}} \\[4pt]
-1, & \text{if } W_i \leq \tau_{\text{low}} \\[4pt]
0, &\text{otherwise}
\end{cases}
\end{equation}
where \( I(i) \) is the indicator function and \( \alpha = \frac{\mu_s \cdot \sigma_s}{\mu_s + \sigma_s} \) modulates the adjustment strength by jointly considering the dispersion and mean of the confidence scores. A more concentrated distribution reduces \( \alpha \), lessening adjustment, while higher mean confidence increases it.

The student model is then trained to match this refined distribution. The KRD loss is the KL-divergence between the student's predicted similarity distribution and the refined teacher distribution:
\begin{equation}
\mathcal{L}_{KRD} = \sum_{i=1}^{K} KL\left( \text{log($\sigma$}\left({s_i^s}\right)) \| \text{$\sigma$}\left({\tilde{s}_i^t}\right) \right)
\end{equation}
where $\sigma$ is the softmax function. This refinement paradigm ensures a more robust and efficient knowledge transfer.

\subsection{Learning and Inference}
In the exploration branch of the student model, we jointly employ InfoNCE loss and Triplet Ranking loss to enhance the discriminative ability of the learned representations. The total loss for this branch is denoted as $L_E$ \cite{dong2023dual}\cite{faghri2017vse++}\cite{zhang2021video}.

For the inheritance branch, we similarly adopt InfoNCE loss and Triplet Ranking loss to enhance consistency with the teacher model's representations, resulting in a loss term  $L_I$. In addition,  we incorporate the knowledge distillation loss $L_{\text{KRD}}$, refined through the KRD module, to further strengthen knowledge transfer from the teacher. Following the dynamic distillation strategy proposed in DL-DKD, an exponentially decaying weight  \( w \) is applied to gradually reduce the influence of the teacher as training progresses. The overall training objective is formulated as:
\begin{equation}
L = L_E + L_I + w \cdot L_{\text{KRD}}
\end{equation}

During inference, only the student model is used for retrieval. Given a video-text pair \((Q, V)\), we independently compute similarity scores using representations from the inheritance and exploration branches, denoted as \({S}_I(Q, V)\) and \({S}_E(Q, V)\), respectively. These scores are then fused via weighted aggregation to obtain the final similarity:
\begin{equation}
\text{Sim}(Q, V) = \delta \cdot {S}_E(Q, V) + (1 - \delta) \cdot {S}_I(Q, V)
\end{equation}
where \(\delta \in [0, 1]\) is a tunable coefficient that adjusts the relative contribution of each branch. Given a textual query \(Q\), all candidate videos \(V\) are ranked based on the computed similarity scores \(\text{Sim}(Q, V)\).

\section{Experiment}
\label{sec:experiment}

\begingroup
\setlength{\aboverulesep}{0pt}
\setlength{\belowrulesep}{0pt}
\setlength{\abovetopsep}{0pt}
\setlength{\belowbottomsep}{0pt}
\begin{table*}[t]
\centering
\caption{Performance comparison on ActivityNet Captions, TVR and Performance on Different M/V Intervals of TVR (only existing open-source models). All metrics are reported as percentages(\%). R@K denotes Recall at rank K.}
\label{tab:merged_results}
\small
\setlength{\tabcolsep}{3pt}
\begin{tabular}{@{}lccccc|ccccc|ccc@{}}
\toprule
\rowcolor{gray!15}
\textbf{Model} & \multicolumn{5}{c|}{\textbf{ActivityNet Captions}} & \multicolumn{5}{c|}{\textbf{TVR}} & \multicolumn{3}{c}{\textbf{TVR:  M/V  Intervals}} \\
\cmidrule(lr){2-6} \cmidrule(lr){7-11} \cmidrule(lr){12-14}
\rowcolor{gray!15}
& \textbf{R@1} & \textbf{R@5} & \textbf{R@10} & \textbf{R@100} & \textbf{SumR} & \textbf{R@1} & \textbf{R@5} & \textbf{R@10} & \textbf{R@100} & \textbf{SumR} & \textbf{(0, 0.2]} & \textbf{(0.2, 0.4]} & \textbf{(0.4, 1]} \\
\midrule
CE \cite{liu2019use}& 5.5 & 19.1 & 29.9 & 71.1 & 125.6 & 3.7 & 12.8 & 20.1 & 64.5 & 101.1 & 104.1 & 149.5 & 165.5 \\
W2VV++ \cite{li2019w2vv++}& 5.4 & 18.7 & 29.7 & 68.8 & 122.6 & 5.0 & 14.7 & 21.7 & 61.8 & 103.2 & 99.0 & 140.7 & 161.3 \\
VSE++ \cite{faghri2017vse++}& 4.9 & 17.7 & 28.2 & 67.1 & 117.9 & 7.5 & 19.9 & 27.7 & 66.0 & 121.1 & 119.1 & 109.8 & 133.5 \\
DE \cite{dong2019dual}& 5.6 & 18.8 & 29.4 & 67.8 & 121.7 & 7.6 & 20.1 & 28.1 & 67.6 & 123.4 & 123.2 & 139.7 & 161.3 \\
DE++ \cite{dong2021dual}& 5.3 & 18.4 & 29.2 & 68.0 & 121.0 & 8.8 & 21.9 & 30.2 & 67.4 & 128.3 & 124.7 & 131.4 & 149.0 \\
RIVRL \cite{dong2022reading}& 5.2 & 18.0 & 28.2 & 64.4 & 117.8 & 9.4 & 23.4 & 32.2 & 70.6 & 135.6 & 133.0 & 121.1 & 135.1 \\
XML \cite{lei2020tvr}& 5.3 & 19.4 & 30.6 & 73.1 & 128.4 & 10.0 & 26.5 & 37.3 & 81.3 & 155.1 & 156.7 & 151.5 & 151.0 \\
ReLoCLNet \cite{zhang2021video}& 5.7 & 18.9 & 30.0 & 72.0 & 126.6 & 10.7 & 28.1 & 38.1 & 80.3 & 157.1 & 157.7 & 153.1 & 156.7 \\
MS-SL \cite{dong2022partially}& 7.1 & 22.5 & 34.7 & 75.8 & 140.1 & 13.5 & 32.1 & 43.4 & 83.4 & 172.4 & 169.6 & 163.4 & 175.8 \\
PEAN \cite{jiang2023progressive}& 7.4 & 23.0 & 35.5 & 75.9 & 141.8 & 13.5 & 32.8 & 44.1 & 83.9 & 174.2 & / & / & / \\
T-D3N \cite{cheng2024transferable}& 7.3 & 23.8 & 36.0 & 76.6 & 145.1 & 13.8 & 33.8 & 45.0 & 83.9 & 176.5 & / & / & / \\
DL-DKD \cite{dong2023dual}& 8.0 & 25.0 & 37.5 & 77.1 & 147.6 & 14.4 & 34.9 & 45.8 & 84.9 & 179.9 & 179.8 & 175.8 & 179.9 \\
LH \cite{fang2024linguistic}& 7.4 & 23.5 & 35.8 & 75.8 & 142.4 & 13.2 & 33.2 & 44.4 & 85.5 & 176.3 & / & / & /\\
GMMFormer \cite{wang2024gmmformer}& 8.3 & 24.9 & 36.7 & 76.1 & 146.0 & 13.9 & 33.3 & 44.5 & 84.9 & 176.6 & 176.2 & 172.8 & 177.4 \\
BGM-Net \cite{yin2024exploiting}& 7.2 & 23.8 & 36.0 & 76.9 & 143.9 & 14.1 & 34.7 & 45.9 & 85.2 & 179.9 & / &/ & / \\
\rowcolor{highlightcolor}
\textbf{KDC-Net} & \textbf{8.1} & \textbf{25.3} & \textbf{38.0} & \textbf{77.5} & \textbf{148.8} & \textbf{15.4} & \textbf{36.3} & \textbf{47.6} & \textbf{85.6} & \textbf{184.9} & \textbf{184.4} & \textbf{178.5} & \textbf{183.9} \\
\bottomrule
\end{tabular}
\end{table*}
\endgroup

\begingroup
\setlength{\aboverulesep}{0pt}
\setlength{\belowrulesep}{0pt}
\setlength{\abovetopsep}{0pt}
\setlength{\belowbottomsep}{0pt}
\setlength{\tabcolsep}{3pt}
\begin{table}[t]
\centering
\caption{Ablation studies of each component on TVR}
\label{tab:ablation-components}
\begin{tabular}{@{}ccc|rrrrr@{}}
\toprule
\rowcolor{gray!15}
\multicolumn{3}{c}{\textbf{Modules}} & \multicolumn{5}{c}{\textbf{Metrics}} \\
\cmidrule(lr){1-3} \cmidrule(lr){4-8}
\rowcolor{gray!15}
\textbf{HSA} & \textbf{DTA} & \textbf{KRD} & \textbf{R@1} & \textbf{R@5} & \textbf{R@10} & \textbf{R@100} & \textbf{SumR} \\
\midrule
  &  &  & 14.4 & 34.1 & 45.0 & 84.3 & 177.8 \\
 $\surd$ &  &  & 14.2 & 34.5 & 45.7 & 84.6 & 179.0 \\
 $\surd$ & $\surd$ &  & \textbf{15.5} & 36.2 & 46.6 & 85.3 & 183.6 \\
 $\surd$ &  & $\surd$ & 14.9 & 35.4 & 46.1 & 85.4 & 181.8 \\
  & $\surd$ &  & 15.2 & 35.8 & 46.0 & 85.1 & 182.1 \\
  &  & $\surd$ & 15.0 & 35.0 & 45.4 & 85.1 & 180.6 \\
  & $\surd$ & $\surd$ & 15.3 & 35.9 & 46.9 & 85.4 & 183.7 \\
\rowcolor{highlightcolor}
 $\surd$ & $\surd$ & $\surd$ & 15.4 & \textbf{36.3} & \textbf{47.6} & \textbf{85.6} & \textbf{184.9} \\
\bottomrule
\end{tabular}
\end{table}
\endgroup

\subsection{Datasets}
We conduct experiments on two standard benchmarks: TVR and ActivityNet Captions. TVR \cite{lei2020tvr} includes 21.8K untrimmed videos with moment-level annotations; we use 17.4K for training and 2.2K for evaluation. ActivityNet \cite{krishna2017dense} Captions contains 20K annotated YouTube videos. We follow the standard split protocol from  for fair comparison.

\subsection{Evaluation Metrics}
We report retrieval performance using Recall@K (R@K, K \(\in \{1, 5, 10, 100\} \)) and SumR. R@K measures the proportion of correct results in the top-K, and SumR reflects overall retrieval quality. All metrics are reported as percentages, with higher values indicating better performance.

\subsection{Implementation Details}
For student model, we follow MS-SL \cite{dong2022partially} to encode queyr-video features, and uses the hidden dimension of 384. For CLIP teacher model, we adopt Vision Transformer ViT-B/32, and encode queries and videos to 512 dimension features. On TVR, we trained for 100 epochs with an initial learning rate of 0.0002. On ActivityNet Captions, we used an initial learning rate of 0.00025 and trained for 50 epochs. All experiments are performed on a single NVIDIA RTX 4090 GPU.

\subsection{Comparison with the State-of-the-Art}
\Cref{tab:merged_results} present comparisons between KDC-Net and existing PRVR methods. On TVR, KDC-Net achieves highest SumR of 184.9, outperforming DL-DKD by +5.0 points and GMMFormer by +8.3 points, while setting a best R@1 record of 15.4\%. These results demonstrate the KDC-Net’s strength in high-precision retrieval across diverse queries. On the more challenging ActivityNet Captions, KDC-Net also achieves the best SumR of 148.5 and consistently surpasses prior methods on R@5 and R@10. Notably, it improves over the PRVR baseline MS-SL by 6.0\% in SumR, underscoring its effectiveness in complex video retrieval scenarios.

To enable finer-grained analysis, we follow the comparison criteria proposed by MS-SL to divide the TVR dataset into three intervals based on the moment-to-video (M/V) ratio of each query, which measures the proportion of relevant content within a video. Lower M/V values indicate shorter target segments amid more irrelevant content, thus posing greater retrieval challenges. This evaluation helps assess robustness under varying query difficulties and content sparsity. Detailed results across three M/V intervals are shown in \cref{tab:merged_results}, where our method clearly outperforms the compared approaches.

\begin{figure*}[ht]
    \centering
    \includegraphics[width=1.0\linewidth]{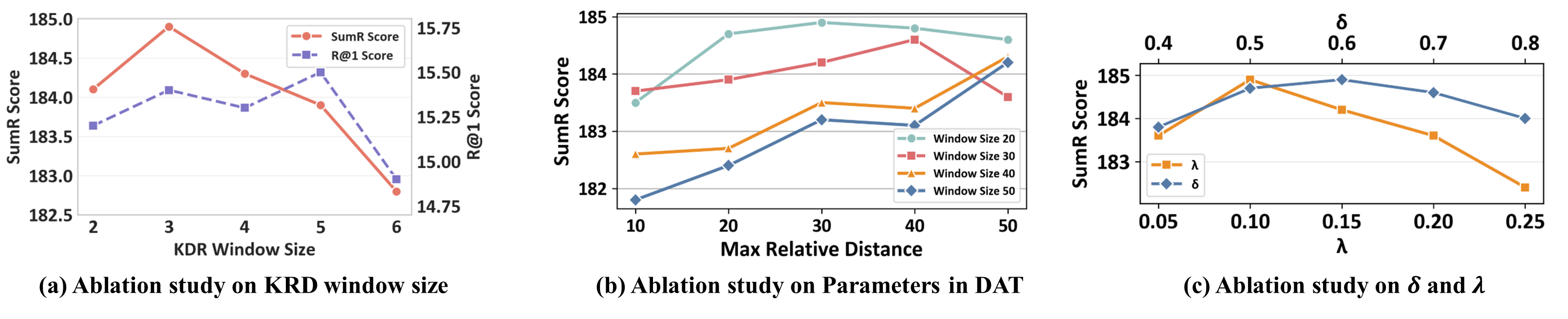} 
    \caption{Ablation studies: (a) KRD window size ablation; (b) DTA parameters ablation; (c) $\delta$ and $\lambda$ parameters ablation.}
    \label{fig:all_ablation_studies}
\end{figure*}

\subsection{Ablation Study}
\subsubsection{Analysis on Components}
We conduct ablation studies to assess the impact of each component in KDC-Net, as summarized in \cref{tab:ablation-components}. Removing any of the three modules results in a noticeable performance drop from the full model to the baseline, validating components' contributions. Among them, HSA is essential for fine-grained cross-modal alignment, while DTA improves temporal relation modeling and reduces redundancy. KRD enhances teacher supervision and distillation effectiveness.

\begingroup
\begin{table*}[t]
\caption{Ablation studies on each branch}
\label{tab:ablation-branch}
\centering
\setlength{\aboverulesep}{0pt}
\setlength{\belowrulesep}{0pt}
\setlength{\abovetopsep}{0pt}
\setlength{\belowbottomsep}{0pt}
\setlength{\tabcolsep}{3pt}
\begin{tabular}{@{}cc|rrrrr|rrrrr@{}}
\toprule
\rowcolor{gray!15}
\multicolumn{2}{c}{\textbf{Branch}} & \multicolumn{5}{c}{\textbf{ActivityNet}} & \multicolumn{5}{c}{\textbf{TVR}} \\
\cmidrule(lr){1-2} \cmidrule(lr){3-7} \cmidrule(lr){8-12}
\rowcolor{gray!15}
\textbf{Inheritance} & \textbf{Exploration} & \textbf{R@1} & \textbf{R@5} & \textbf{R@10} & \textbf{R@100} & \textbf{SumR} & \textbf{R@1} & \textbf{R@5} & \textbf{R@10} & \textbf{R@100} & \textbf{SumR} \\
\midrule
& $\surd$ & 6.2 & 20.5 & 33.5 & 73.9 & 134.1 & 12.7 & 32.1 & 42.4 & 83.9 & 171.1 \\
$\surd$ &  & 7.8 & 24.5 & 36.5 & 76.8 & 145.6 & 11.8 & 30.2 & 41.0 & 82.6 & 165.6 \\
\rowcolor{highlightcolor}
$\surd$ & $\surd$ & \textbf{8.1} & \textbf{25.3} & \textbf{38.0} & \textbf{77.5} & \textbf{148.8} & \textbf{15.4} & \textbf{36.3} & \textbf{47.6} & \textbf{85.6} & \textbf{184.9} \\
\bottomrule
\end{tabular}
\end{table*}
\endgroup

\subsubsection{Analysis on DTA}
We investigate the impact of key hyperparameters in DTA by varying the maximum relative position and window size as shown in \cref{fig:all_ablation_studies} (b). Results reveal that performance improves when the two are configured in a balanced manner. Specifically, increasing either parameter leads to performance gains up to a point, after which the improvement plateaus, reflecting a trade-off between receptive field expansion and local attention precision. The best SumR score is achieved at a maximum relative position of 30 and window size of 20. Notably, the model performs best when their ratio is around 1.5, suggesting that a slightly broader relative position range than the attention window effectively enhances spatial bias modeling.

We further study the impact of the scale parameter \(\lambda\) in the PN module, as shown in  \cref{fig:all_ablation_studies} (c). Performance peaks at \(\lambda = 0.1\), indicating optimal feature separation. Notably, exceeding this value causes a sharper decline in performance compared to lower values, suggesting that over-regularization is more detrimental than under-regularization. These findings validate the PN module’s design in effectively balancing discriminability and information preservation.

\subsubsection{Analysis on KRD}
We conduct an ablation study on TVR to assess the impact of sliding window size in the KRD module, varying it from 2 to 6. As shown in  \cref{fig:all_ablation_studies} (a), the best performance is achieved with a window size of 3. A smaller window limits inter-frame interaction, hindering effective learning, while a larger window introduces noise from irrelevant frames, degrading alignment quality. Window sizes of 3–5 yield consistently strong results, highlighting the importance of modeling fine-grained, localized frame intervals for effective retrieval.

\subsubsection{Analysis on Parameters and Branches}
In order to investigate the parameter $\delta$ used in the Inference stage to balance the weights of exploration branch and inheritance branch, we conducted ablation experiments as shown in  \cref{fig:all_ablation_studies} (c). The results show that the best performance is achieved when $\delta$ is 0.1.
The experiments shown in \cref{tab:ablation-branch} demonstrate that the dual-branch learning strategy adopted by KDC-Net is effective. Simply using the teacher signal from the inherited branch for learning or independently learning through the exploration branch alone cannot achieve higher results.

\section{Conclusion}
\label{sec:conclusion}

We propose KDC-Net, a novel method for PRVR that achieves new state-of-the-art results through three key innovations: adaptive hierarchical semantic modeling, dynamic temporal visual encoding, and dynamic refined knowledge distillation. Extensive experiments on TVR and ActivityNet Captions demonstrate consistent improvements over strong baselines, especially under low M/V ratio scenarios. The results of comprehensive comparison and ablation experiments have confirmed the ability of KDC-Net in bridging the semantic gap between complex text queries and untrimmed video content, laying a solid foundation for future PRVR research.






\bibliographystyle{sec/IEEEbib}
\bibliography{sec/main}











\end{document}